%% file: main.tex
\documentclass{article}
\usepackage{iclr2026_conference,times}

\input{math_commands.tex}

\usepackage{graphicx}
\usepackage{hyperref}
\usepackage{booktabs}
\usepackage{amssymb}
\usepackage{pifont}
\usepackage{amsmath}
\usepackage{array}
\usepackage{algorithm}
\usepackage{algorithmic}
\usepackage{xcolor}
\usepackage{makecell}
\usepackage{subcaption}
\usepackage{enumitem}
\usepackage[most,skins,breakable]{tcolorbox}
\usepackage{colortbl}
\usepackage{multirow}

\definecolor{superlightgray}{gray}{0.95}
\definecolor{borderblack}{gray}{0.1}

\newenvironment{promptfigure}[3][superlightgray]{%
  \begin{figure*}[h]
    \centering
    \def\promptCaption{#2}%
    \def\promptLabel{#3}%
    \begin{tcolorbox}[
        width=\textwidth,
        colback=#1,
        colframe=borderblack,
        boxrule=0.5pt,
        arc=3mm,
        left=3mm, right=3mm, top=3mm, bottom=3mm,
        fontupper=\small\raggedright,
        halign=flush left,
        parskip=0.5\baselineskip
      ]
      \setlist[itemize]{leftmargin=1.5em, nosep}
      \setlist[enumerate]{leftmargin=1.5em, nosep}
    }{%
    \end{tcolorbox}
    \caption{\promptCaption}
    \label{\promptLabel}
  \end{figure*}
}

\title{Experiential Reflective Learning for Self-Improving LLM Agents}

\author{Marc-Antoine Allard\thanks{Equal contribution.}\quad
  Arnaud Teinturier\footnotemark[1]\quad
  Victor Xing\footnotemark[1]\quad
  Gautier Viaud \\[0.5ex]
  Illuin Technology\\
  \texttt{\{marc-antoine.allard,victor.xing\}@illuin.tech} \\
}

\iclrfinalcopy
\begin{document}

\maketitle

\begin{abstract}
  Recent advances in large language models (LLMs) have enabled the development of autonomous agents capable of complex reasoning and multi-step problem solving. However, these agents struggle to adapt to specialized environments and do not leverage past interactions, approaching each new task from scratch regardless of their accumulated experience. We introduce \textbf{Experiential Reflective Learning} (ERL), a simple self-improvement framework that enables rapid environment adaptation through experiential learning. ERL reflects on task trajectories and outcomes to generate heuristics, capturing actionable lessons that transfer across tasks. At test time, relevant heuristics are retrieved based on the current task and injected into the agent's context to guide execution. On the Gaia2 benchmark, ERL improves success rate by 7.8\% over a ReAct baseline, with large gains in task completion reliability, and outperforms prior experiential learning methods. Through systematic ablations, we find that selective retrieval is essential and that heuristics provide more transferable abstractions than few-shot trajectory prompting. These results demonstrate that reflecting on single-attempt experiences to extract transferable heuristics enables effective agent self-improvement.
\end{abstract}

\section{Introduction}
Agentic systems powered by large language models (LLMs) are increasingly deployed for complex tasks requiring multi-step planning, reasoning, and tool use~\citep{luo2025mcpuniversebenchmarkinglargelanguage, yao2025taubench, wei2025browsecompsimplechallengingbenchmark}. However, general-purpose agents often fail to adapt to new environments with unfamiliar tools and domain-specific conventions~\citep{chen2026groundedtesttimeadaptationllm}. Fine-tuning can enable such adaptation but is resource-intensive, infeasible for closed-source models, and does not support continuous learning. These limitations have motivated research into experiential memory systems that enable parameter-free improvement through accumulated experience~\citep{hu2026memoryageaiagents}.

Recent work has explored how to make LLM agents learn from accumulated experience without parameter updates~\citep{cai2025flexcontinuousagentevolution, fang2026mempexploringagentprocedural, sarukkai2025selfgeneratedincontextexamplesimprove, ouyang2025reasoningbankscalingagentselfevolving, fu2024autoguide, zhao2024expel}. ExpeL~\citep{zhao2024expel} extracts reusable insights by comparing successful and failed trajectories, using Reflexion to retry each task until success. Extracted insights are then concatenated into every test prompt regardless of task relevance, an approach that scales poorly as experience accumulates. AutoGuide~\citep{fu2024autoguide} creates context-aware guidelines on offline training tasks by contrasting paired trajectories with different outcomes. At test time, it performs context identification and guideline retrieval at every agent turn, incurring substantial overhead, and provides no guidance when the current state fails to match any stored context. Both methods require multiple rollouts per task to construct contrastive trajectory pairs, an assumption that breaks down in practical agent deployment where tasks cannot be retried.

In this work, we propose \textbf{Experiential Reflective Learning} (ERL), an experiential memory framework designed for efficient self-improvement in new environments. ERL builds a pool of reusable heuristics that capture effective strategies and failure modes, generated by reflecting on past experience trajectories and their outcomes. Then, for each new task, an LLM scores stored heuristics for relevance and injects the top candidates into the agent's context, providing task-specific guidance. ERL extracts heuristics from single-attempt trajectories, enabling efficient adaptation without curated training sets or repeated execution. Moreover, ERL heuristics preserve granular trajectory details that existing methods lose through cross-task aggregation.

On the Search and Execution splits of Gaia2~\citep{froger2025arescalingagentenvironments}, ERL achieves an overall success rate of 56.1\%, an improvement of +7.8\% over a ReAct baseline and outperforming ExpeL and AutoGuide. We carry out experiments highlighting that heuristic retrieval plays a critical role in improving performance and reliability, lessons from failures and successes benefit different task types, and heuristics provide more transferable abstractions than raw agent trajectories.

\begin{figure}[t]
  \centering
  \includegraphics[width=0.8\linewidth]{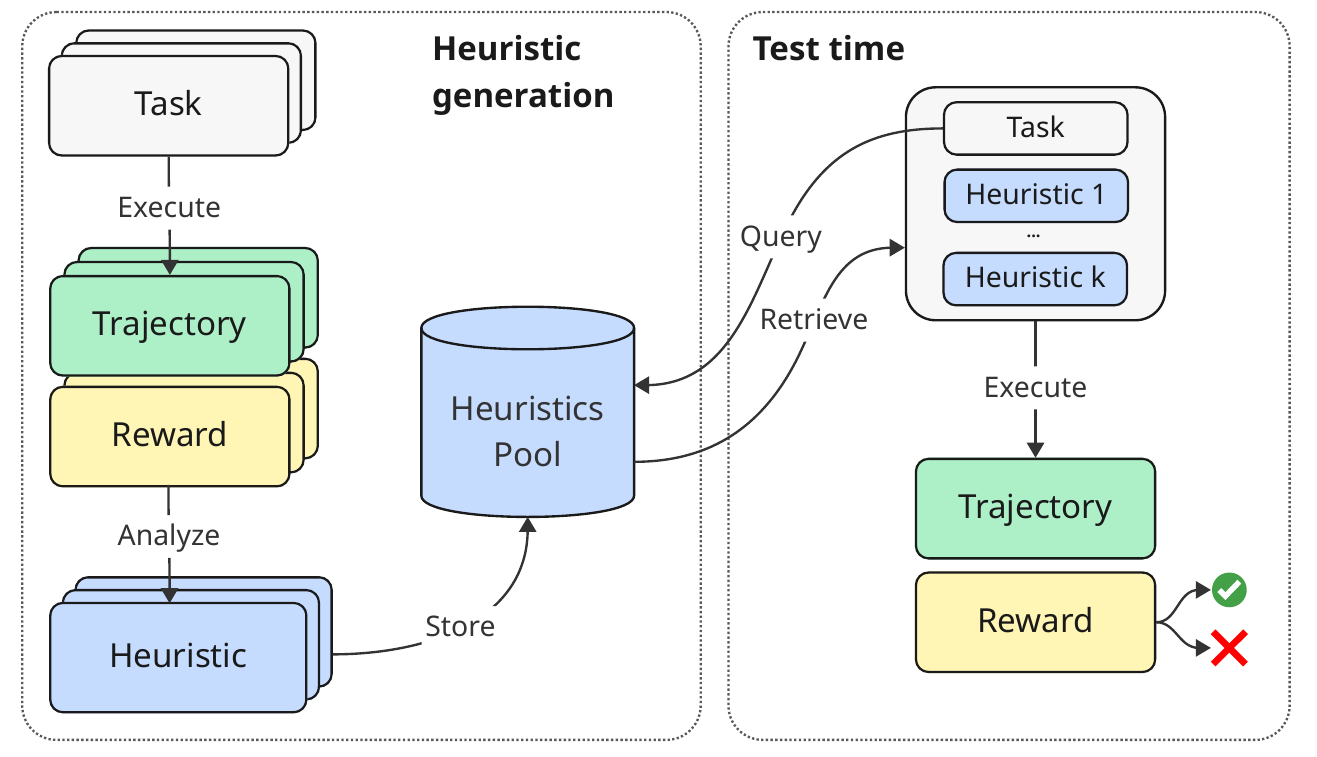}
  \caption{The ERL framework. During experience accumulation (left), the agent reflects on task outcomes to generate heuristics stored in a persistent pool. At test time (right), relevant heuristics are retrieved and injected into context for new tasks.}
  \label{fig:schema_erl}
\end{figure}

\section{Methods : Experiential Reflective Learning}

ERL consists of two components (Figure~\ref{fig:schema_erl}): heuristic generation from task experience, and retrieval-augmented execution for new tasks.

\paragraph{Heuristic generation.}
We consider an agent operating in an environment that provides binary success/failure feedback upon task completion (we analyze performance when outcome signals are unavailable in Appendix~\ref{appendix:result}). As the agent executes tasks, it accumulates experiences consisting of the task description, the execution trajectory (reasoning steps, tool calls, and outputs), and the outcome signal. After each task, the agent reflects on this experience to generate a structured heuristic containing: (1) an analysis identifying what led to success or failure, and (2) a learned guideline with explicit trigger conditions and recommended actions (e.g., "When sending emails to calendar attendees, first resolve names to email addresses via the Contacts tool before calling the email API"). These heuristics are stored in a persistent pool.

\paragraph{Retrieval-augmented execution.}
When facing a new task, the agent retrieves relevant heuristics from the pool to guide its execution and improve from past experiences. An LLM analyzes the new task, decomposes it into potential sub-tasks and action steps, and scores each stored heuristic for relevance. Selection is made based on the similarity between the stored and current task descriptions, the diversity of experiences to cover a range of potential lessons, and the informativeness of the guideline content. The top-$k$ heuristics are then injected into the agent's system prompt, ensuring that the agent receives task-specific advice rather than being overwhelmed with the full heuristic pool. Appendix~\ref{app:examples} provides an example heuristic and a trajectory where the agent explicitly references the guidance it was given, and Appendix~\ref{app:prompts} contains the prompts for heuristic generation and retrieval.

\section{Experiments}

\begin{figure}[!b]
  \centering
  \begin{minipage}[b]{0.45\linewidth}
    \centering
    \includegraphics[width=\linewidth]{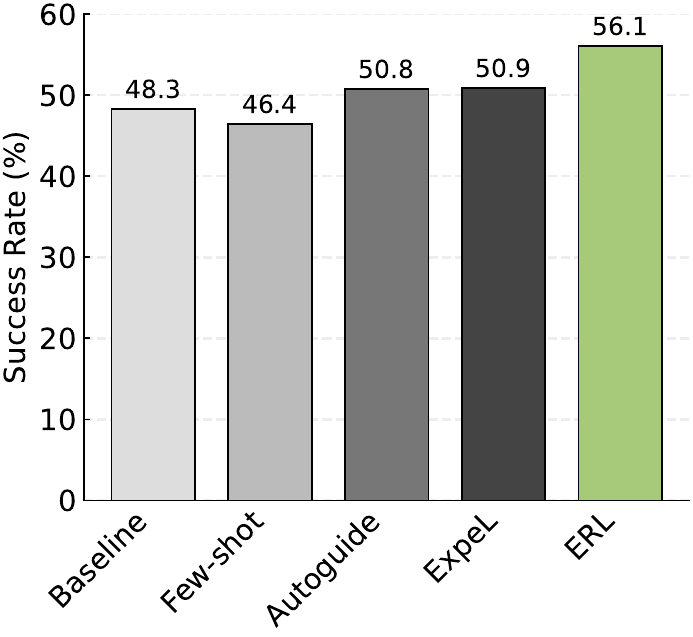}
    \subcaption{Averaged across Search and Execution splits.}
    \label{fig:main_results}
  \end{minipage}
  \hfill
  \begin{minipage}[b]{0.52\linewidth}
    \centering
    \setlength{\tabcolsep}{4pt}
    \begin{tabular*}{\linewidth}{@{\extracolsep{\fill}}l ccc}
      \toprule
      & \multicolumn{3}{c}{\textbf{Success Rate (\%)}} \\
      \cmidrule(lr){2-4}
      \textbf{Method} & Exec. & Search & \cellcolor{blue!10}Overall \\
      \midrule
      Baseline & 43.1 & 53.6 & \cellcolor{blue!10}48.3 \\
      Few-shot & 41.7 & 51.2 & \cellcolor{blue!10}46.4 \\
      ExpeL & 45.8 & 56.0 & \cellcolor{blue!10}50.9 \\
      AutoGuide & 39.6 & 61.9 & \cellcolor{blue!10}50.8 \\
      \midrule
      \textbf{ERL (Ours)} & \underline{51.4} & 60.7 & \cellcolor{blue!10}\underline{56.1} \\
      \textit{Ablations:} & & & \cellcolor{blue!10} \\
      \quad no retrieval & 44.4 & \underline{63.1} & \cellcolor{blue!10}53.8 \\
      \quad embedding retrieval & 50.7 & 56.0 & \cellcolor{blue!10}53.3 \\
      \quad only failures & 50.0 & \textbf{67.9} & \cellcolor{blue!10}\textbf{58.9} \\
      \quad only successes & \textbf{52.1} & 47.6 & \cellcolor{blue!10}49.9 \\
      \bottomrule
    \end{tabular*}
    \subcaption{Per-split results and ablations.}
    \label{tab:detail_results}
  \end{minipage}
  \caption{Success rates on the Gaia2 test universes (3 runs). \textbf{Bold}: best; \underline{underline}: second best.}
  \label{fig:main_combined}
\end{figure}

\paragraph{Experimental setup.} We evaluate ERL against existing approaches on Gaia2~\citep{froger2025arescalingagentenvironments}, a recent benchmark that assesses long-horizon agentic capabilities within a simulated mobile environment containing 12 applications and 101 tools. We focus on the Search and Execution splits which contain tasks involving information retrieval and multi-step execution capabilities.

We use the default ReAct agent scaffold of Gaia2 as the baseline method in our evaluations. ERL augments this baseline by injecting retrieved heuristics into the agent's system prompt before execution, requiring no modifications to the core ReAct loop. Unless otherwise noted, we evaluate ERL with $k=20$ heuristics retrieved by an LLM (see Appendix~\ref{Retrieval} for discussions on the retrieval method and choice of $k$). We use GPT-5-mini as the agent backbone for all methods.

A key feature of Gaia2 is its organization into \textit{universes}: isolated data partitions that expose identical applications and tools but contain completely disjoint information (e.g., different contacts, files, and calendar events). This structure enables evaluation without knowledge contamination: we accumulate heuristics on 8 universes (112 execution / 132 search tasks) and evaluate on 2 held-out test universes (48 / 28 tasks). This setup mirrors real-world deployment where agents typically operate within a fixed set of tools while encountering continuously evolving data. Additional evaluation details are available in Appendix~\ref{appendix:environment}.

\paragraph{Main results.} Figure~\ref{fig:main_combined} summarizes performance across methods. ERL achieves a 56.1\% overall success rate, an improvement of +7.8\% over the ReAct baseline and +5.2\% over the strongest prior method (ExpeL at 50.9\%). Gains are consistent across task types: +8.3\% on Execution and +7.1\% on Search relative to the baseline. Prior methods exhibit uneven performance across splits, as AutoGuide achieves strong Search results (61.9\%) but underperforms on Execution (39.6\%, below baseline), while ExpeL shows modest improvements on both. Few-shot prompting with raw trajectory demonstrations also fails to improve over the baseline (46.4\%).

\begin{figure}[t]
  \centering
  \begin{subfigure}{\linewidth}
    \centering
    \includegraphics[width=\linewidth]{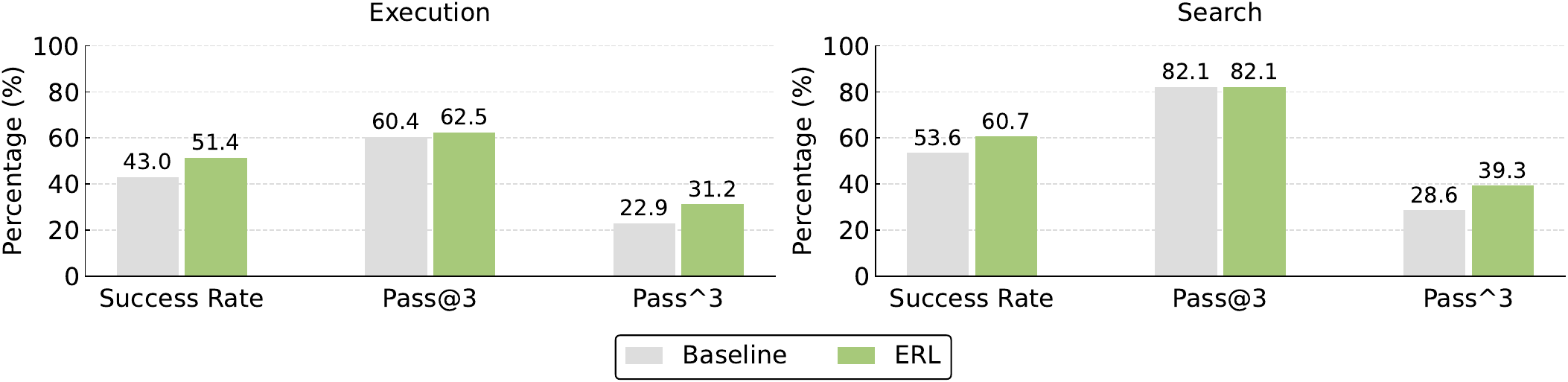}
  \end{subfigure}
  \caption{Success rate, pass@3 and pass\char`\^3 comparison between baseline and ERL}
  \label{fig:sr_pass_scores}
\end{figure}

\paragraph{ERL improves agent reliability.}
Figure~\ref{fig:sr_pass_scores} compares pass@3 and pass\char`\^3 between the baseline and ERL. pass@3 measures the fraction of scenarios where the agent succeeds at least once across three runs, capturing whether a task is within the agent's capability. pass\char`\^3 requires success on all three runs, measuring whether the agent can reliably complete a task~\citep{chen2021evaluatinglargelanguagemodels,yao2025taubench}. ERL yields substantial gains on pass\char`\^3 (+8.3\% in Execution and +10.6\% in Search), indicating more stable performance across runs. Improvements on pass@3 are comparatively smaller, suggesting only marginal gains in newly solved scenarios.

\begin{figure}[t]
  \centering
  \includegraphics[width=0.8\linewidth]{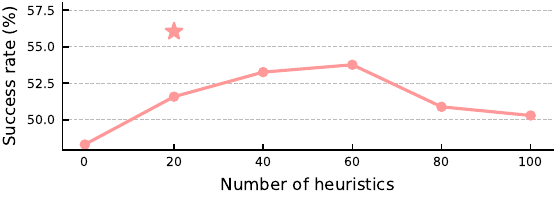}
  \caption{Success rate vs. number of randomly selected heuristics. ERL with LLM-based retrieval of $k=20$ heuristics ($\bigstar$) outperforms all random configurations.}
  \label{fig:nb_self_reflections}
\end{figure}

\paragraph{Heuristics generalize better than trajectories.}
A simple approach to in-context learning from experiences is to append raw trajectories  as few-shot demonstrations. However, this approach fails to improve performance (-1.9\% vs. baseline), suggesting that raw trajectories lack the actionable insights necessary for the agent to improve on new scenarios. On the other hand, heuristics provide distilled strategic principles that generalize across tasks and are more context-efficient. We show that this advantage holds across token budgets, as heuristics consistently outperform trajectories even when controlling for context length (Appendix~\ref{appendix:result}).

\paragraph{Retrieval quality matters more than quantity.}
ERL retrieves task-relevant heuristics using an LLM-based selection mechanism. To validate the value of this choice, we compare against random selection. Figure~\ref{fig:nb_self_reflections} shows average success rate as a function of the number of randomly selected heuristics. Performance exhibits a non-monotonic pattern, peaking around 40–60 heuristics before degrading with excessive inclusion. This motivates the need for a selection mechanism that identifies the most relevant heuristics from an extensive pool, rather than relying on quantity alone. LLM-based retrieval with $k=20$ achieves 56.1\% overall, outperforming both embedding-based retrieval using Qwen3-Embedding-0.6B~\citep{zhang2025qwen3embeddingadvancingtext} (53.3\%, Table~\ref{tab:detail_results}) and the best random selection configuration (53.8\%), confirming that retrieval quality matters more than heuristic quantity.

\paragraph{Failure heuristics favor Search; success heuristics favor Execution.}
Prior methods explore learning specifically from past failures or successes~\citep{sarukkai2025selfgeneratedincontextexamplesimprove,zhao2024expel,shinn2023reflexion}. We examine whether the outcome of source experiences affects heuristic quality. Table~\ref{tab:detail_results} shows that failure-derived heuristics substantially outperform success-derived ones overall. The effect is split-dependent (Table~\ref{tab:detail_results}): failure heuristics excel on Search (+14.3\% over baseline) by providing negative constraints that prune ineffective strategies, whereas success heuristics work best on Execution (+9.0\%) by reinforcing proven action sequences. While failure-only retrieval yields the highest overall score (58.9\%), this reflects strong Search performance at the expense of Execution tasks. In practical deployment scenarios where the task distribution is unknown, retrieving from both sources offers a reliable compromise.

\section{Conclusion}
We introduce Experiential Reflective Learning (ERL), a framework enabling LLM agents to improve through accumulated experience. By distilling trajectories into reusable heuristics and retrieving relevant guidance at test time, ERL outperforms prior experiential learning methods and improves task completion reliability on Gaia2. Our analysis reveals that (1) heuristics transfer better than raw trajectories, (2) LLM-based retrieval outperforms random and embedding-based selection, and (3) learning from failures versus successes has a variable impact depending on the task type. Future work could bootstrap heuristic accumulation via synthetic task generation or address challenges in scaling heuristic pools, such as resolving conflicting guidelines and maintaining retrieval quality.

\section*{Acknowledgments}

This work was carried out within the framework of the LIAGORA ”LabCom”, a joint laboratory supported by the French National Research Agency (ANR) and established between ILLUIN Technology and the MICS laboratory of CentraleSupelec.

\bibliographystyle{iclr2026_conference}
\bibliography{iclr2026_conference}

\newpage

\appendix

\section{$\tau^2$-Bench Evaluation}
While Gaia2 evaluates agents in a single-control setting, real-world deployment often requires agents
to coordinate with users who actively participate in task resolution. We therefore evaluate ERL on
$\tau^2$-bench~\citep{barres2025tau2benchevaluatingconversationalagents}, a benchmark spanning three customer service domains (Airline, Retail, and Telecom).

Table~\ref{tab:tau2bench} shows that ERL improves overall success rate over the baseline (0.380 vs.\ 0.367), with gains on Airline and Retail but a slight drop on Telecom. On Airline and Retail, ERL improves pass\char`\^3 while pass@3 decreases. This corroborates our findings on Gaia2 where ERL mainly improved consistency through pass\char`\^3 gains with modest pass@3 changes.

Telecom follows the opposite trend on every metric: success rate drops, pass\char`\^3 falls to zero, yet pass@3 improves (0.625 vs.\ 0.575). We hypothesize that two design differences between Telecom and the other domains may explain this divergence. First, Telecom tasks are composed combinatorially from atomic subtasks, yielding a large configuration space that reduces the likelihood of any heuristic matching the specific subtask composition of a test task. Second, Telecom is a dual-control domain where the agent must guide a user who has their own tools, a randomly assigned persona, and variable device state. This makes the interaction less predictable than in single-control domains like Airline and Retail. As a result, a retrieved heuristic may improve the agent's diagnostic strategy on some trials, but fail on others when the user interaction unfolds differently.

Overall, results on $\tau^2$-bench corroborate the Gaia2 findings in single-control domains, while the Telecom results highlight limitations around capturing user coordination strategies that could be an interesting direction for future work.

\begin{table}[h]
  \centering
  \setlength{\tabcolsep}{6pt}
  \begin{tabular}{l l cccc}
    \toprule
    &  & Airline & Retail & Telecom & \cellcolor{blue!10}Overall \\
    \midrule
    \multirow{2}{*}{\textbf{Success Rate}}
    & Baseline & 36.7 & 43.3 & \textbf{30.0} & \cellcolor{blue!10}36.7 \\
    & ERL & \textbf{38.3} & \textbf{47.5} & 28.3 & \cellcolor{blue!10}\textbf{38.0} \\
    \midrule
    \multirow{2}{*}{\textbf{pass\char`\^3}}
    & Baseline & 20.0 & 12.5 & \textbf{10.0} & \cellcolor{blue!10}\textbf{14.2} \\
    & ERL & \textbf{25.0} & \textbf{17.5} & 0.0 & \cellcolor{blue!10}\textbf{14.2} \\
    \midrule
    \multirow{2}{*}{\textbf{pass@3}}
    & Baseline & \textbf{55.0} & \textbf{80.0} & 57.5 & \cellcolor{blue!10}\textbf{64.2} \\
    & ERL & 50.0 & 75.0 & \textbf{62.5} & \cellcolor{blue!10}62.5 \\
    \bottomrule
  \end{tabular}
  \caption{Per-domain results (in \%) on $\tau^2$-bench. Success rates are averaged over 3 trials per task. Evaluation is performed on the test split of each domain. ERL uses heuristics generated from the training split. \textbf{Bold}: best per group.}
  \label{tab:tau2bench}
\end{table}

\section{Gaia2 evaluation setup} \label{appendix:environment}

We conduct experiments on the Agents Research Environments (ARE) platform~\citep{froger2025arescalingagentenvironments}, which provides infrastructure for running and evaluating agents on the Gaia2 benchmark. ARE provides a default ReAct agentic scaffold which we use as our baseline. Each scenario defines a task along with a ground-truth trajectory, enabling automatic verification of task completion.

Test universes (22 and 25) were randomly selected from the available pool. As detailed in Table~\ref{tab:model_selection}, we select GPT-5-mini to maintain an optimal balance between reasoning performance, inference latency, and computational cost.

\begin{table}[h]
  \centering
  \renewcommand{\arraystretch}{1.3}
  \begin{tabular}{ll}
    \toprule
    \textbf{Role} & \textbf{LLM} \\
    \midrule
    Agent & gpt-5-mini-2025-08-07 \\
    ARE LLM Judge & gpt-5-mini-2025-08-07 \\
    Heuristic generation & gpt-5-mini-2025-08-07 \\
    Heuristic retrieval & gpt-5.2-2025-12-11 \\
    \bottomrule
  \end{tabular}
  \caption{LLM model selection}
  \label{tab:model_selection}
\end{table}

\paragraph{ExpeL~\citep{zhao2024expel} implementation.}

We sample a subset of training tasks matching the test set distribution (48 execution, 28 search). Following the original implementation, we collect success and failure experiences by running the Reflexion~\citep{shinn2023reflexion} algorithm with GPT-5-mini for up to 3 retries. For insights extraction, we use a batch size of $L=3$ successful trajectories. During inference, we retrieve 3 task-relevant few-shot examples using the Qwen3-Embedding-0.6B embedding model.

\paragraph{AutoGuide~\citep{fu2024autoguide} implementation.}
To generate context-aware guidelines, we run Reflexion with GPT-5-mini for up to 3 retries on all training tasks, and run the guideline generation process on scenarios with contrastive successful/failed trajectories. At test-time, the top 3 relevant guidelines are contextually retrieved and fed into the agent context after each \textit{observation} ReAct turn. We use GPT-5-mini for guideline selection, extraction and context identification.

\section{Additional experiments} \label{appendix:result}

\subsection{Token-matched comparison with a few-shot baseline}
\begin{figure}[h]
  \centering
  \includegraphics[width=\linewidth]{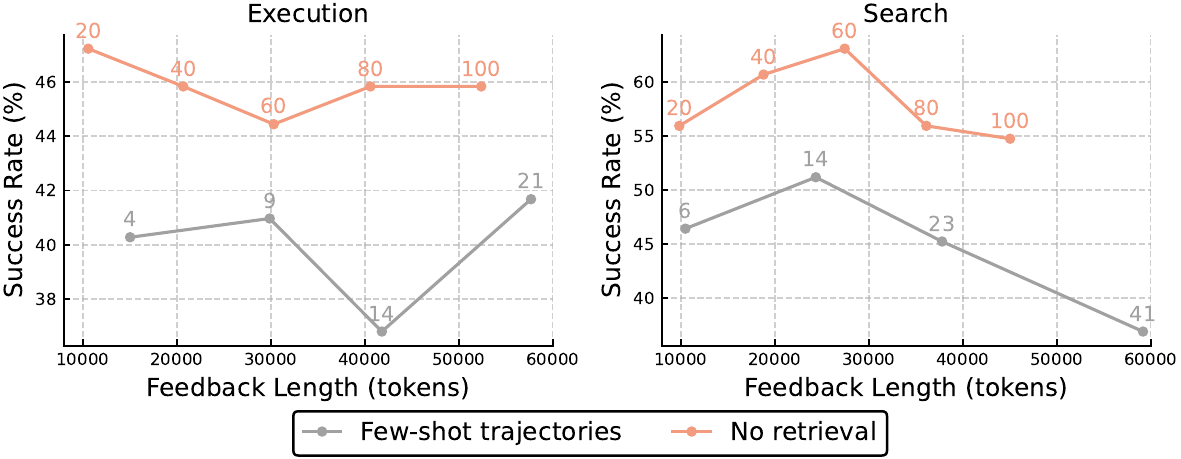}
  \caption{Success Rate over feedback token count for heuristic feedback and few-shot trajectories.}
  \label{fig:sr_vs_few_shot}
\end{figure}

We investigate whether heuristics outperform few-shot trajectories beyond token efficiency alone. Figure~\ref{fig:sr_vs_few_shot} demonstrates that for any number of experiences, using heuristics leads to higher success rates than raw trajectories. This holds when comparing the same number of experiences provided in-context as either heuristics or raw trajectories: heuristics yield a +5.5\% gain in Execution at approximately 20 scenarios and a +23.8\% gain in Search at 40 scenarios, where raw trajectories approach a regime where long-context performance starts to degrade. This confirms that distilling experiences into heuristics provides a superior learning signal compared to raw trajectory accumulation, while enabling substantially more experiences to fit within a fixed context budget.

\subsection{Heuristic retrieval} \label{Retrieval}

To determine the optimal retrieval approach and number of heuristics $k$ to retrieve, we evaluate two methods: embedding-based and LLM-based retrieval. We limit our experiments to at most $k = 20$ heuristics, as this represents a practical upper bound: retrieving too many heuristics can be detrimental in complex scenarios, and the LLM-based retrieval approach becomes inefficient for $k > 20$. The experimental setup follows the general environment configuration described in Section~\ref{appendix:environment}.

\paragraph{Embedding retrieval.}
We first evaluate embedding-based retrieval using the Qwen3-Embedding-0.6B model, computing cosine similarity between the test task and all training tasks, then selecting the top-$k$ most similar heuristics. This approach effectively retrieves contextually similar tasks that use the same tools (e.g., a calendar task retrieves all calendar-related training scenarios). However, it does not necessarily prioritize heuristics addressing similar error patterns. Table~\ref{tab:topk_ablation_sr_qwen} shows the best overall performance is achieved with $k=20$, yielding a 53.3\% success rate.

\paragraph{LLM retrieval.}
To account for similarity between the test task and the content of the feedback itself, we evaluate LLM-based retrieval using GPT-5.2 as the ranker. The prompt in Figure~\ref{fig:heuristic_retrieval_prompt} instructs the model to consider not only task similarity but also error patterns and the lessons articulated in each heuristic. Table~\ref{tab:topk_ablation_sr_llm} shows that LLM-based ranking outperforms embedding-based retrieval for most values of $k$, particularly on the Search split, with optimal performance again at $k=20$ with a 56.1\% success rate.

\paragraph{Importance of the reward signal.}
Reward signals may not always be available in real-world agentic tasks. We investigate a configuration where environment validation rewards are unavailable during heuristic generation, requiring the agent to infer the outcome of its trajectory and derive generalizable guidelines from this self-assessment. In this setting, the agent correctly identifies success or failure only 70\% of the time, and the success rate of ERL drops to 51.2\% (-4.8\%). We note that this still exceeds the baseline (48.3\%), indicating that ERL provides value even with imperfect reward signals, though accurate outcome feedback remains important for optimal performance.
\begin{table}[h]
  \centering
  \begin{minipage}[b]{0.48\linewidth}
    \centering
    \setlength{\tabcolsep}{6pt}
    \begin{tabular}{l ccc}
      \toprule
      & \multicolumn{3}{c}{\textbf{Success Rate (\%)}} \\
      \cmidrule(lr){2-4}
      \textbf{$k$} & Execution & Search & \cellcolor{blue!10}Overall \\
      \midrule
      1  & 43.1 & \textbf{57.1} & \cellcolor{blue!10}50.1 \\
      2  & 49.3 & 44.1 & \cellcolor{blue!10}46.7 \\
      3  & 43.8 & \textbf{57.1} & \cellcolor{blue!10}50.5 \\
      4  & 50.0 & \underline{56.0} & \cellcolor{blue!10}\underline{53.0} \\
      5  & \textbf{51.4} & 47.6 & \cellcolor{blue!10}49.5 \\
      10 & 45.8 & \underline{56.0} & \cellcolor{blue!10}50.9 \\
      15 & 47.2 & 51.2 & \cellcolor{blue!10}49.2 \\
      20 & \underline{50.7} & \underline{56.0} & \cellcolor{blue!10}\textbf{53.3} \\
      \bottomrule
    \end{tabular}
    \subcaption{Embedding-based (Qwen3-Embedding-0.6B)}
    \label{tab:topk_ablation_sr_qwen}
  \end{minipage}
  \hfill
  \begin{minipage}[b]{0.48\linewidth}
    \centering
    \setlength{\tabcolsep}{6pt}
    \begin{tabular}{l ccc}
      \toprule
      & \multicolumn{3}{c}{\textbf{Success Rate (\%)}} \\
      \cmidrule(lr){2-4}
      \textbf{$k$} & Execution & Search & \cellcolor{blue!10}Overall \\
      \midrule
      1  & 45.1 & 53.6 & \cellcolor{blue!10}49.4 \\
      3  & 47.2 & \underline{58.3} & \cellcolor{blue!10}52.8 \\
      5  & \underline{49.3} & \underline{58.3} & \cellcolor{blue!10}\underline{53.8} \\
      10 & \textbf{51.4} & 54.8 & \cellcolor{blue!10}53.1 \\
      20 & \textbf{51.4} & \textbf{60.7} & \cellcolor{blue!10}\textbf{56.1} \\
      \bottomrule
    \end{tabular}
    \subcaption{LLM-based (GPT-5.2)}
    \label{tab:topk_ablation_sr_llm}
  \end{minipage}
  \caption{Effect of retrieved heuristics count ($k$) by retrieval method. \textbf{Bold}: best; \underline{underline}: second best.}
  \label{tab:topk_ablation}
\end{table}

\section{Token count and cost analysis}\label{appendix:tokens_cost}

\begin{table}[H]
  \centering
  \small
  \begin{tabular}{llrrrrr}
    \toprule
    & & \multicolumn{2}{c}{\textbf{Tokens (M)}} & & \\
    \cmidrule(lr){3-4}
    \textbf{Configuration} & \textbf{Step} & \textbf{Input \scriptsize\textcolor{gray}{(\% cached)}} & \textbf{Output} & \textbf{Cost (\$)} & \textbf{Avg. Turns} \\
    \midrule
    Baseline & Scenario rollout  & 103.8 {\scriptsize\textcolor{gray}{(82\%)}} & 4.3 & 15.27 & \multirow{1}{*}{16.6} \\
    \midrule
    \multirow{4}{*}{ERL}
    & Heuristic generation & 7.5\,\, {\scriptsize\textcolor{gray}{(8\%)}}    & 0.3 & 2.42  & \multirow{4}{*}{17.6} \\
    & Heuristic retrieval            & 1.0 {\scriptsize\textcolor{gray}{(86\%)}}   & 0.1 & 1.38  & \\
    & Scenario rollout   & 192.0 {\scriptsize\textcolor{gray}{(88\%)}} & 3.9 & 17.60 & \\
    \cmidrule{2-5}
    & \textit{Total}       & \textit{200.5} {\scriptsize\textcolor{gray}{(85\%)}} & \textit{4.3} & \textit{21.40} & \\
    \bottomrule
  \end{tabular}
  \caption{Breakdown of token usage and costs for the evaluation of the baseline and ERL agents. All experiments used the official OpenAI API with flex processing and models listed in Table~\ref{tab:model_selection}.}
  \label{tab:computational_overhead}
\end{table}

We report total cumulative token counts and API costs for one full evaluation run on our Gaia2 test set, comparing the baseline ReAct agent against ERL with LLM retrieval of 20 heuristics per task. Input tokens for scenario rollout nearly double (+85\%), primarily because retrieved heuristics ($\sim$20k tokens) are appended to the base system prompt ($\sim$12k tokens) at every turn, while the number of turns per scenario remains roughly constant (16.6 vs.\ 17.6). Thanks to prompt caching, which absorbs a larger share of input tokens under ERL (88\% vs. 82\%), the cost increase for scenario rollout alone is a moderate +15\%. Accounting for heuristic generation and retrieval, ERL incurs a 40\% overall increase in API costs. Future work could explore more compact heuristic representations to preserve agent self-adaptation at lower computational overhead.

We do not measure latency, as it depends heavily on the model provider's API load. However, we note the additional LLM call involved in ERL to retrieve relevant heuristics is marginal for most Gaia2 tasks that span dozens of turns, but may be non-negligible for shorter tasks.

\section{Iterative ERL}

We also evaluate an iterative variant of ERL where the agent retrieves heuristics from the pool as it grows, rather than accumulating all heuristics first. Tasks are processed in batches; within each batch, the agent retrieves guidance from the current heuristic pool before execution, then adds newly generated heuristics to the pool (Algorithm~\ref{alg:iterative-reflective-learning-llm}). This creates a cumulative learning effect: early batches establish foundational knowledge from naive failures, while later batches benefit from this guidance and potentially encounter more nuanced errors. Comparing these two approaches tests whether learning from progressively improving trajectories offers advantages over learning from independent executions.

\begin{figure}[h]
  \centering
  \begin{minipage}{0.5\textwidth}
    \begin{algorithm}[H]
      \caption{Iterative ERL}
      \label{alg:iterative-reflective-learning-llm}
      \begin{algorithmic}[1]
        \STATE \textbf{Input:} $N$ batches, size $B$, retrieval count $k$
        \STATE \textbf{Init:} Pool $\mathcal{P} \leftarrow \varnothing$
        \FOR{$i = 1$ \textbf{to} $N$}
        \STATE Sample batch $\mathcal{B}_i$
        \FOR{\textbf{each} $x \in \mathcal{B}_i$}
        \STATE $C \leftarrow \textsc{Retrieve}(\mathcal{P}, x, k)$
        \STATE $\tau \leftarrow \textsc{Execute}(x, C)$
        \STATE $r \leftarrow \textsc{Reward}(\tau)$
        \STATE $h \leftarrow \textsc{Analyze}(x, \tau, r)$
        \STATE $\mathcal{P} \leftarrow \mathcal{P} \cup \{(x, r, h)\}$
        \ENDFOR
        \ENDFOR
        \STATE \textbf{Return:} $\mathcal{P}$
      \end{algorithmic}
    \end{algorithm}
  \end{minipage}
\end{figure}

\begin{table}[h]
  \centering
  \setlength{\tabcolsep}{4pt}
  \begin{tabular}{l cc c ccc}
    \toprule
    & \multicolumn{2}{c}{\textbf{Source Tasks (\%)}} & & \multicolumn{3}{c}{\textbf{Test Tasks (\%)}} \\
    \cmidrule(lr){2-3} \cmidrule(lr){5-7}
    \textbf{Method} & Execution & Search & & Execution & Search & \cellcolor{blue!10}Overall \\
    \midrule
    Baseline & — & — & & 43.1 & 53.6 & \cellcolor{blue!10}48.3 \\
    \midrule
    ERL & 42.0 & 53.8 & & \textbf{51.4} & \textbf{60.7} & \cellcolor{blue!10}\textbf{56.1} \\
    Iterative ERL & \textbf{44.6} & \textbf{59.9} & & 46.5 & 54.8 & \cellcolor{blue!10}50.7 \\
    \makecell[l]{Iterative ERL (only failures)} & 41.1 & 56.8 & & \underline{47.9} & \textbf{60.7} & \cellcolor{blue!10}\underline{54.3} \\
    \bottomrule
  \end{tabular}
  \caption{Success rates on source and test tasks for ERL variants. Iterative ERL achieves higher source task performance but lower test generalization. \textbf{Bold}: best; \underline{underline}: second best.}
  \label{tab:erl_variants}
\end{table}

In Table~\ref{tab:erl_variants}, we report success rates on the test universes and on the source universes on which heuristics are accumulated. The scores reveal that iterative ERL achieves higher success rates than the standard version during accumulation but lower test performance (-5.4\%). Several factors may explain this generalization gap. First, as iterative guidance improves performance, the agent may encounter fewer and narrower failure modes. The resulting heuristics could miss the breadth of naive mistakes that new tasks are likely to trigger. Second, heuristics from guided trajectories may be less transferable, as they reflect behavior shaped by prior guidance rather than independent discovery by the agent.

\section{Heuristic and trajectory example}
\label{app:examples}

Figure~\ref{fig:heuristic_example} shows a sample heuristic. Heuristics follow a set structure with an analysis of why the agent succeeded or failed and a guideline with generalizable takeaways. We observe that guidelines often make reference to specific tools that the agent can call, and clarify idiosyncratic behaviors and errors that can steer future behavior.

Figure~\ref{fig:traj_example} shows an abridged trajectory where the agent explicitly refers to a heuristic during the completion of a new task. The heuristic, derived from a past failure where the agent forgot to delete an original event when rescheduling, prescribes a \textit{safe reschedule} procedure: create the replacement event first, then delete the original. Although the new task involves replacing events rather than rescheduling, the agent recognizes the structural similarity and repeatedly applies the guideline (underlined), ensuring each original event is deleted after its replacement is created.

\section{Heuristic generation and retrieval prompts}
\label{app:prompts}

The heuristic generation prompt is provided in Figure~\ref{fig:heuristic_prompt}. It takes as input a task, the trajectory of the agent, and the associated reward (success or failure). We guide the heuristics to emphasize constructive insights, enabling the agent to avoid repeating past mistakes and to build upon effective reasoning patterns when forming future reflections.

Heuristic retrieval at test time is performed by an LLM with structured outputs using the prompt in Figure~\ref{fig:heuristic_retrieval_prompt}. This version was selected after a human comparison with other variants containing more detailed rubrics and different scoring ranges. To handle retrieval a large pool of heuristics, future work could explore a two-stage process combining an embedding model to collect a large set of candidate heuristics, and the LLM to rerank them and select the top $k$.

\begin{promptfigure}{Heuristic example}
  {fig:heuristic_example}
  Scenario ID: scenario\_universe\_27\_9aqrhh\\\smallskip

  \textbf{Task:} Please order me an American Girl Pink Owl Bedroom Decor Girls Craft Kit that is of intermediate skill level. Also, can you reschedule my first upcoming dinner with colleagues to 10/20/2024 instead from 6 pm to 8 pm? Send an email to the attendees to inform them of the updated event. Additionally, remove all saved properties from my saved list that cost exactly \$2000.\\\smallskip

  \textbf{Reward:} failure\\\smallskip

  \begin{enumerate}
    \item \textbf{Analysis:}
      \begin{itemize}
        \item \textbf{Breakpoint:} I attempted to call \texttt{Emails\_\_send\_email} using attendee names (``Sergei Kuznetsov'', ``Viktor Kuznetsov'') instead of valid email addresses. The email tool requires real email addresses and threw an ``Invalid email address'' error. Although I recovered by looking up contacts and resending successfully, the initial incorrect tool call is the direct cause of the failure (invalid tool input).\\\smallskip
      \end{itemize}

    \item \textbf{Learned Guideline:}
      \begin{itemize}
        \item \textit{Trigger:} When I need to send an email and my input is attendee names (from calendar events, meeting attendees, or other name-only lists).
        \item \textit{Action:} Resolve and validate recipient addresses before calling the email tool:
          \begin{itemize}
            \item \textbf{Step 1:} For each attendee name, attempt \texttt{Contacts\_\_get\_contact} (exact) or \texttt{Contacts\_\_search\_contacts} (fuzzy) to retrieve an email address.
            \item \textbf{Step 2:} If multiple contacts match, pick the best match deterministically (exact name match, same city, or most recent contact) and log the choice; if ambiguous or no match, ask the user for clarification or the correct email.
            \item \textbf{Step 3:} Validate the resulting addresses have correct email format (contains `@' and a domain) before calling \texttt{Emails\_\_send\_email}.
            \item \textbf{Step 4:} If \texttt{Emails\_\_send\_email} returns an ``Invalid email address'' error, parse the error to identify which recipient(s) failed, remove or correct only those entries, and retry rather than reissuing the original invalid request.\\\smallskip
          \end{itemize}
      \end{itemize}
  \end{enumerate}

  \textbf{Rationale:} Pre-resolving and validating recipients prevents invalid tool inputs, reduces retries, and keeps execution robust and auditable when sending messages on the user's behalf.
\end{promptfigure}

\begin{promptfigure}{Example trajectory with explicit reference to heuristics}{fig:traj_example}

  \subsubsection*{Task}

  \begin{quote}
    I want to dedicate next month to young wine enthusiasts, so delete all events
    scheduled for next month that contain ``Wine tasting'' in the title and have
    one or more attendee over 45 years old. Schedule new events titled ``Wine
    Awakening'' to replace them with no attendees and the same timings.
    [\ldots]
  \end{quote}

  \subsubsection*{Injected Heuristic (excerpt)}

  \begin{quote}
    \textbf{Analysis:} Root cause of failure: I created a new calendar event
    instead of actually ``rescheduling'' the existing event---I left the original
    event intact.

    \textbf{Learned Guideline:}\\
    \emph{Trigger:} When instructed to ``reschedule'', ``move'', or ``change the
    date/time'' of an existing calendar event.\\
    \emph{Action:} I must explicitly locate the original \texttt{event\_id}, then
    either (A)~modify it using an edit API if available, or (B)~if no edit API
    exists, create the new event \textbf{and delete the original event}. After
    doing either, verify the change by querying the calendar to confirm the
    original event is removed and the new event exists.
  \end{quote}

  \subsubsection*{Agent Trajectory (excerpt)}

  \begin{description}

    \item[\normalfont\textit{[\ldots the agent retrieves calendar events and contact
      records\ldots]}]~

    \item[Step 3.] \textbf{Thought:} I can see a November~5 ``Wine Tasting
      Event'' that includes Vasile Munteanu (age~70), so it meets the criterion.
      \underline{Per the safe reschedule guideline}, I should create the
      replacement event first with the same timings and no attendees, then delete
      the original.\\
      \textbf{Action:} \texttt{Calendar\_\_add\_calendar\_event}
      \{title: ``Wine Awakening'', start: 2024-11-05 18:00, end: 2024-11-05
      20:00, attendees: []\}

    \item[Step 4.] \textbf{Thought:} I created the replacement.
      \underline{Per the safe reschedule procedure,} I should now delete the
      original event.\\
      \textbf{Action:} \texttt{Calendar\_\_delete\_calendar\_event}
      \{event\_id: ca87b8\ldots\}

    \item[\normalfont\textit{[\ldots the agent looks up the ages of attendees on
      the November~8 event\ldots]}]~

    \item[Step 7.] \textbf{Thought:} The Nov~8 event includes August Blackwood
      (age~79), so it must be replaced. \underline{Following the safe-reschedule
      procedure,} I'll create a new ``Wine Awakening'' event with the same details
      but no attendees.\\
      \textbf{Action:} \texttt{Calendar\_\_add\_calendar\_event}
      \{title: ``Wine Awakening'', start: 2024-11-08 18:00, end: 2024-11-08
      20:00, attendees: []\}

    \item[Step 8.] \textbf{Thought:} \underline{To complete the safe reschedule
      (create new event then remove the original),} I will now delete the original
      event.\\
      \textbf{Action:} \texttt{Calendar\_\_delete\_calendar\_event}
      \{event\_id: eb831b\ldots\}

    \item[\normalfont\textit{[\ldots the same pattern repeats for Nov~12 and
      Nov~20\ldots]}]~

  \end{description}

\end{promptfigure}

\begin{promptfigure}{Heuristic generation prompt}{fig:heuristic_prompt}
  You are an AI agent engaging in self-reflection. You analyze your own trajectories to extract generalizable principles and guidelines that improve your future task execution across diverse scenarios. Your feedback is concise, strategic, and applicable beyond the specific task analyzed.\\\smallskip

  You are an intelligent agent operating in the ARE environment. You have just completed a task trajectory.
  Your goal is to perform a ``Post-Mortem Analysis'' to extract transferable lessons that will improve your reliability and reasoning on future tasks.\\\smallskip

  \textbf{Context:}
  \begin{itemize}
    \item Task Description: \{task\_info\}
    \item Outcome: \{validation\_info\}
    \item Trajectory: \{trajectory\_text\}\\\smallskip
  \end{itemize}

  \textbf{Instructions:}
  Analyze your performance based on the Outcome.\\\smallskip

  \textbf{IF FAILURE:}
  \begin{enumerate}
    \item \textbf{Pinpoint the Breakpoint:} Locate exactly where the logic failed or the tool was misused.
    \item \textbf{Derive a Correction Rule:} Create a specific guideline that prevents this class of error.
      \begin{itemize}
        \item \textit{Bad:} ``I should be more careful with files.''
        \item \textit{Good:} ``When a file path is unknown, I must use \texttt{ls -R} or \texttt{find} before attempting to \texttt{read\_file} to avoid PathNotFound errors.''\\\smallskip
      \end{itemize}
  \end{enumerate}

  \textbf{IF SUCCESS:}
  \begin{enumerate}
    \item \textbf{Identify the ``Winning Move'':} What specific decision or tool usage made this efficient?
    \item \textbf{Derive a Best Practice:} Create a rule to reinforce this behavior.
      \begin{itemize}
        \item \textit{Example:} ``When asked to summarize data, aggregating it into a temporary file before processing is more robust than reading chunks.''\\\smallskip
      \end{itemize}
  \end{enumerate}

  \textbf{Output Format:}
  Construct your response to be appended to your future memory.

  \begin{enumerate}
    \item \textbf{Analysis:} (Briefly explain the cause of success or failure in this specific trace)
    \item \textbf{Learned Guideline:} (A concise, imperative rule for future tasks. Focus on Tool Usage, Verification Loops, or Reasoning Steps.)
      \begin{itemize}
        \item \textit{Trigger:} [When/If I encounter Situation X...]
        \item \textit{Action:} [I must explicitly do Y...]\\\smallskip
      \end{itemize}
  \end{enumerate}

  Reflect now:
\end{promptfigure}

\begin{promptfigure}{Heuristic retrieval prompt}{fig:heuristic_retrieval_prompt}
  \subsection*{Role}
  You are an expert assistant specialized in selecting the most relevant heuristics to help an AI agent improve its performance on a given task. Heuristics are derived from past experiences (successful or failed) and contain a task description, an analysis of task completion, and learned guidelines.

  \subsection*{Goal}
  Your goal is to analyze and decompose the provided task into potential sub-tasks and action steps. Then, from the list of heuristics, identify the TOP \{k\} most relevant heuristics.

  \subsection*{Input}
  You will be provided with:\\\smallskip
  \begin{enumerate}
    \item A task description that the agent is going to perform.
    \item A list of past experiences in the form of heuristics from previous tasks. Composed of:
      \begin{itemize}
        \item Scenario ID
        \item Task description
        \item Reward (success or failure)
        \item Heuristic text
      \end{itemize}
  \end{enumerate}

  \subsection*{Output}
  Your output should be a list of the TOP \{k\} scenario IDs of the most relevant heuristics that can help the agent improve its performance on the given task. You must also return a score between 0 and 100 and a brief justification for this score, based on the scoring criteria below.

  \subsection*{Heuristic Selection}
  When selecting the most relevant heuristics, consider the following criteria:\\\smallskip
  \begin{itemize}
    \item \textbf{Similarity of the task descriptions:} Look for heuristics where the task closely matches or relates to the current task.
    \item \textbf{Diversity of experiences:} Aim to include heuristics that cover a range of scenarios and challenges, providing a broad perspective.
    \item \textbf{Informative content:} Prioritize heuristics with a ``Learned Guideline:'' section that offer detailed insights, lessons learned, and actionable advice.
  \end{itemize}

  \subsection*{Output Format}
  Your output should be a JSON object with the TOP \{k\} scenario IDs. For each ID, first provide a brief justification (1--2 sentences) of the score, then give a score (0--100):

  \begin{tcolorbox}[colback=white, boxrule=0.2pt, arc=1mm, fontupper=\small\ttfamily]
    \{ \\
      \quad "ID number 1": ["rationale", score], \\
      \quad "ID number 2": ["rationale", score], \\
      \quad ... \\
      \quad "ID number k": ["rationale", score] \\
    \}
  \end{tcolorbox}

  \textbf{LIST OF HEURISTICS:} \\
  \texttt{\{heuristics\_list\}}\\\smallskip

  \textbf{TASK TO COMPLETE:} \\
  \texttt{\{task\}}
\end{promptfigure}

\end{document}

%% file: math_commands.tex
\usepackage{amsmath,amsfonts,bm}

\def\eqref#1{equation~\ref{#1}}

\def\1{\bm{1}}

\DeclareMathAlphabet{\mathsfit}{\encodingdefault}{\sfdefault}{m}{sl}
\SetMathAlphabet{\mathsfit}{bold}{\encodingdefault}{\sfdefault}{bx}{n}

